\documentclass[conference,a4paper]{IEEEtran}
\usepackage{graphicx}
\usepackage{multirow}
\usepackage{stfloats}
\usepackage{amsmath}
\usepackage{subcaption}
\usepackage{color}
\usepackage{amssymb}
\begin{document}
% \title{Mitigating Unintentional Facial Identity Learning in Deepfake Detection with Face Recognizers}
\title{FRIDAY: Mitigating Unintentional Facial Identity in Deepfake Detectors Guided by Facial Recognizers}

\author{
\IEEEauthorblockN{
Younghun Kim\textsuperscript{1},
Myung-Joon Kwon\textsuperscript{2},
Wonjun Lee\textsuperscript{2}, and
Changick Kim\textsuperscript{1,2}}
\IEEEauthorblockA{
\textsuperscript{1}Graduate School of Green Growth and Sustainability,
KAIST,
Daejeon, South Korea}
\IEEEauthorblockA{
\textsuperscript{2}School of Electrical Engineering, KAIST, Daejeon, South Korea\\
Email: \{younghun1664, kwon19, dpenguin, changick\}@kaist.ac.kr}
}

% \author{
% \IEEEauthorblockN{Younghun Kim}
% \IEEEauthorblockA{Graduate School of Green Growth \\and Sustainability \\
% KAIST, Republic of Korea\\
% Email: younghun1664@kaist.ac.kr}
% \and
% \IEEEauthorblockN{Myungjun Kwon}
% \IEEEauthorblockA{Twentieth Century Fox\\
% Email: homer@thesimpsons.com}
% \and
% \IEEEauthorblockN{James Kirk\\ and Montgomery Scott}
% \IEEEauthorblockA{Starfleet Academy\\
% San Francisco, California 96678--2391\\
% Telephone: (800) 555--1212\\
% Fax: (888) 555--1212}
% \and
% \IEEEauthorblockN{James Kirk\\ and Montgomery Scott}
% \IEEEauthorblockA{Starfleet Academy\\
% San Francisco, California 96678--2391\\
% Telephone: (800) 555--1212\\
% Fax: (888) 555--1212}
% }

\maketitle

\begin{abstract}
% 하지만 딥페이크 탐지 모델들은 자신이 학습한 in-domain에서는 성능이 좋지만 새로운 기법으로 생성된 cross-domain에서는 성능이 매우 좋지 않습니다.

% 딥페이크는 사회적으로 잘못된 정보 전달 또는 개개인의 인

% Deepfakes cause considerable harm to society by enabling the spread of misinformation and potentially damaging the reputations of individuals. 

Previous Deepfake detection methods perform well within their training domains, but their effectiveness diminishes significantly with new synthesis techniques. Recent studies have revealed that detection models make decision boundaries based on facial identity instead of synthetic artifacts, leading to poor cross-domain performance. To address this issue, we propose FRIDAY, a novel training method that attenuates facial identity utilizing a face recognizer. To be specific, we first train a face recognizer using the same backbone as the Deepfake detector. We then freeze the recognizer and use it during the detector's training to mitigate facial identity information. This is achieved by feeding input images into both the recognizer and the detector, then minimizing the similarity of their feature embeddings using our Facial Identity Attenuating loss. This process encourages the detector to produce embeddings distinct from the recognizer, effectively attenuating facial identity. Comprehensive experiments demonstrate that our approach significantly improves detection performance on both in-domain and cross-domain datasets.

\begin{IEEEkeywords}
Deepfake Detection, Unintentional Facial Identity, Face Recognition, Image Forensics
\end{IEEEkeywords}

% Deepfakes present a substantial challenge to society. While Deepfake detection models exhibit high performance within the domains they were trained on, their effectiveness significantly diminishes when applied to new, previously unseen techniques. Recent studies have revealed that some facial features are not entirely eliminated during training and persist through Deepfake synthesis. This causes detection models to unintentionally make boundary decisions based on facial identity, leading to poor cross-domain performance. We propose FRIDAY, a novel training method to mitigate this issue. Specifically, we first train a face recognizer using the same backbone as the Deepfake detector. After training, we freeze the recognizer and use it during the training of the detector to separate facial identity information. This is achieved by feeding input images into both the recognizer and the detector, and then comparing their feature embeddings using cosine similarity loss. This process encourages the detector to produce distinct embeddings from the recognizer, effectively separating face identity. Comprehensive experimental results show that our approach significantly improves detection performance on both in-domain and cross-domain datasets.
\end{abstract}

\section{Introduction}
In recent years, the advent of deep learning techniques has given rise to sophisticated face forgery methods \cite{DeepFakes, Face2Face, FaceSwap, FaceShifter} for creating highly realistic synthetic videos, commonly known as Deepfakes. These technologies can be used to spread misinformation, manipulate public opinion, commit fraud, and defame individuals \cite{rossler2019faceforensics++}.

% \begin{figure}[]
% \centerline{\includegraphics[scale=0.8]{figures/figure1/figure1 (1).png}}
% \caption{\textit{Unintentional identity learning.} (a) is an example where the identity of the source does not match that of the generated fake image. (b) validation accuracy and (c) training loss are the results of training a model for identification classification by replacing only the classifier head with an identification classification head and freezing the backbone of a model that was initially trained for Deepfake detection.}
% \label{fig:1}
% \end{figure}

\begin{figure}[htbp]
    \centerline{\includegraphics[width=0.5\textwidth]{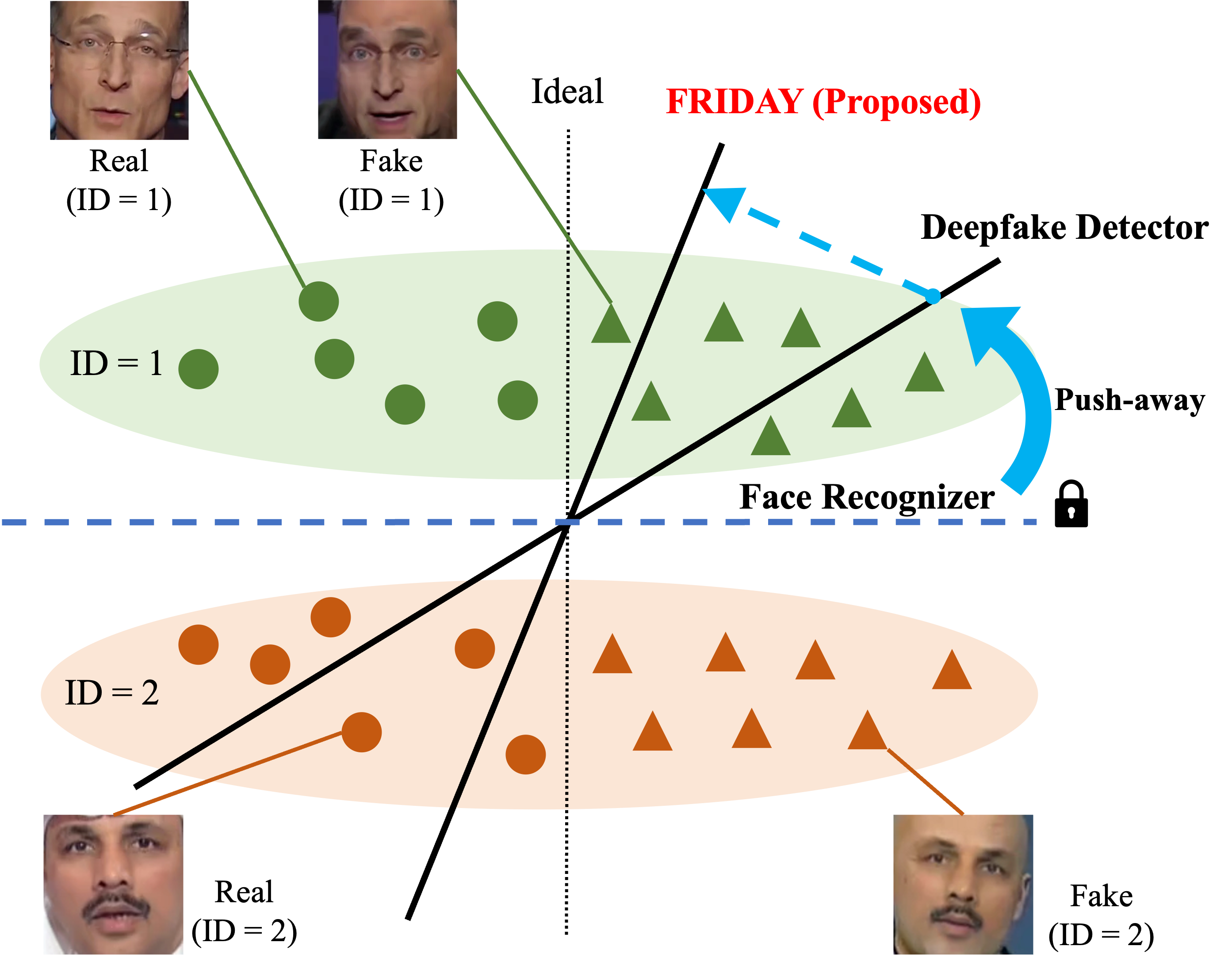}}
        \caption{\textit{Overview of FRIDAY.} During the Deepfake detector training, unintended facial identity learning occurs, causing biased predictions. We address this by pushing the Deepfake detector's decision boundary away from the face recognizer's, leveraging its identity extraction capability.}
    \label{fig:1}
\end{figure}

While previous studies \cite{afchar2018mesonet, nguyen2019capsule} exhibit strong performance on the in-domain datasets, their performance significantly diminishes when applied to cross-domain datasets. To address this generalization problem, several approaches have been proposed: focusing on intra-frame inconsistency \cite{noise_based, uia}, mimicking common defects in Deepfakes \cite{li2020facexray, li2018exposing, shiohara2022sbi}, and leveraging the facial identity information \cite{implicit_identity_leakage, dong2022celebrities}. 

% One of the primary causes of the generalization problems is that the model unintentionally learns the identity of the face during the training process \cite{implicit_identity_leakage}. This occurs because the binary classifier distinguishes between fake and genuine images based on subtle identity differences that remain from the face synthesis process, causing the model to create incorrect identity-based decision boundaries.

% 모델이 cross-domain 예측을 진행할 때는 어쩔 수 없이 학습한 데이터와 이미지 퀄리티, 조작 기법 등이 달라지면서 유효한 특징들이 크게 줄어듭니다. 그래서 일반화를 위해서는 학습과정에서 모델이 무엇보다도 이미지 결함과 관련된 정보에만 집중하게 하는 것이 중요합니다. 하지만, 최근 연구에 따르면 모델이 학습과정에서 의도치 않게 얼굴 정체성을 학습한다는 것이 발견되었습니다. This is because it is easier for the model to minimize loss by creating decision boundaries based on facial identities instead of Deepfake artifacts. 이렇게 유효한 특징이 줄어드는 것은 in-domain datset에서는 여전히 다른 유효한 특징들이 존재해서 큰 문제가 되지 않지만 cross-domain과 같이 안그래도 유효한 특징이 부족한데 더 부족하게 되어 성능이 하락하게 됩니다. 따라서, 무엇보다 모델이 이런 의미 없는 facial identites를 학습하지 않게 하는게 매우 중요합니다. 

When a model makes predictions on cross-domain datasets, the effective features are significantly reduced due to variations in training data, image quality, and manipulation techniques. For this reason, ensuring that the model focuses on features related to artifacts during the training process is important. However, recent studies have found that the model unintentionally learns facial identities during the training \cite{implicit_identity_leakage}. This occurs because it is easier for the model to minimize loss by creating decision boundaries based on facial identities rather than Deepfake artifacts. While the reduction in effective features is not problematic within an in-domain dataset due to the presence of other useful features, it becomes a significant issue in cross-domain datasets where the already limited effective features are further diminished, leading to a performance drop. Therefore, ensuring that the model does not learn facial identities during the training process is crucial.

In this paper, we propose a novel training method called FRIDAY (\textbf{F}ace \textbf{R}ecognizer \textbf{ID}-\textbf{A}ttenuating Methodolog\textbf{Y}), as shown in Fig.\ref{fig:1}, to solve the unintentional facial identity learning problem in the Deepfake detector. This approach involves training the face recognizer and then leveraging it during the training of the Deepfake detector to attenuate the identity information. Intuitively, the face recognizer focuses on facial identity, making it an ideal tool to address the unintended learning of identity representations by the detector. By minimizing our Facial Identity Attenuating loss between the feature embeddings from the face recognizer and those from the Deepfake detection model, we effectively reduce the facial identity features in the detector. This method forces the Deepfake detector to minimize its reliance on facial identity features. Our primary contributions are as follows:

\begin{itemize}
    \item We propose a novel training method called FRIDAY, which can be applied to many Deepfake detectors to improve their generalization performances.
    \item For the first time, we use a face recognizer in the training process of Deepfake detectors to attenuate unintentional facial identity embedding.
    \item We demonstrate the superior performance of our method through comparisons with existing state-of-the-art models in both in-domain and cross-domain scenarios. 
\end{itemize}

\section{Face Recognizer ID-Attenuating Methodology}
% \begin{figure*}[]
% \centerline{\includegraphics[width=0.5\textwidth]{figures/overview.png}}
% \caption{\textit{Overview of FRIDAY.} }
% \label{fig:2}
% \end{figure*}
\subsection{Motivation}
% \begin{figure}[]
% \centerline{\includegraphics[width=0.45\textwidth]{figures/figure1/epoch_loss.png}}
% \caption{\textit{FRIDAY Phase 1.} This describes the training process for the face recognizer used in Deepfake detector training. The face recognizer is trained using a closed-set and identification learning approach.}
% \label{fig:epoch}
% \end{figure}

\begin{figure}[t!]
    \centering
    \begin{subfigure}[b]{0.48\textwidth}
        \includegraphics[width=\textwidth]{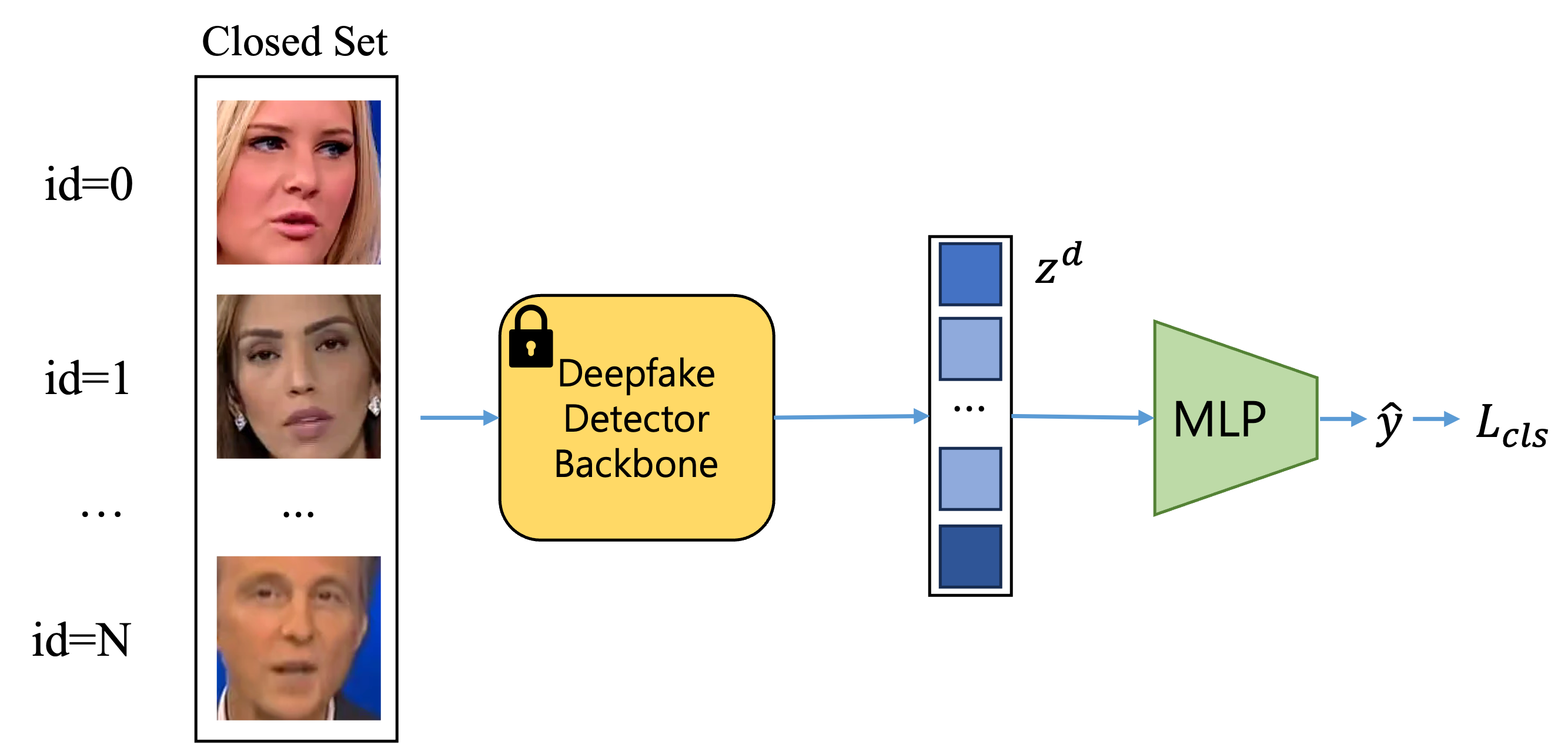}
        \caption{Unintentional identity learning check method.}
        \label{fig:face_identity_a}
    \end{subfigure}
    \centering
    \begin{subfigure}[b]{0.48\textwidth}
        \includegraphics[width=\textwidth]{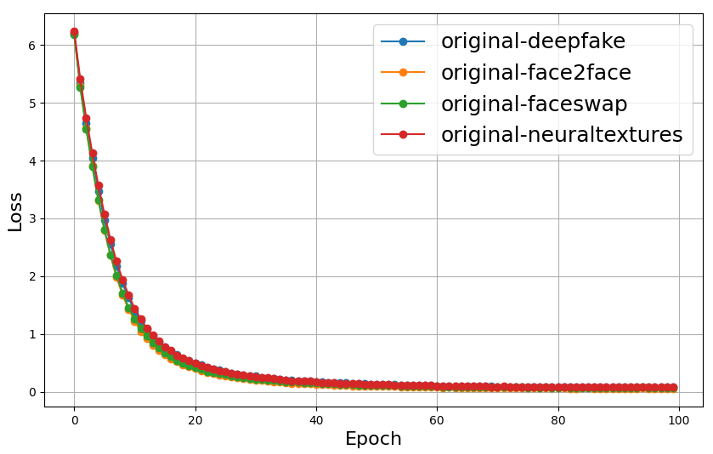}
        \caption{Epoch Loss.}
        \label{fig:face_identity_b}
    \end{subfigure}
    % \hfill
    % \begin{subfigure}[b]{0.23\textwidth}
    %     \includegraphics[width=\textwidth]{figures/figure1/epoch_valid.png}
    %     \caption{Valid Accuracy.}
    %     \label{fig:face_identity_c}
    % \end{subfigure}
    \caption{\textit{Demonstrating how much FRIDAY reduces face identity information.}}
    \label{fig:face_identity}
\end{figure}

After training the Deepfake detector, we freeze it and replace the classifier with a trainable face identification classifier. We then continue face classification training, as shown in Fig.~\ref{fig:face_identity_a}. Interestingly, as shown in Fig.~\ref{fig:face_identity_b}, the model easily converges in all datasets. This convergence indicates that the model inherently contains a substantial amount of face identity features. This phenomenon, where facial identity is implicitly present, adversely affects the model's generalization performance \cite{implicit_identity_leakage}.

Face recognizers utilize a significant amount of facial identity information to distinguish faces. Consequently, we aim to reduce facial identity information in the Deepfake detector by leveraging a face recognizer. This concept results in the FRIDAY learning method, whose overall process is depicted in Fig.~\ref{fig:phase1} and Fig.~\ref{fig:phase2}. The learning process of this method is divided into two phases. In the first phase, we train the face recognizer. In the second phase, we train the Deepfake detector, reducing facial identity information using the face recognizer trained in the first phase. We achieve this reduction by incorporating Facial Identity Attenuating loss between the embeddings of the Deepfake detector and the face recognizer. Although both models share the same structure, they do not share parameters of backbones and classifiers.

\subsection{Phase 1: Face Recognizer Training}
% \begin{figure}[]
% \centerline{\includegraphics[width=0.45\textwidth]{figures/phase1.png}}
% \caption{\textit{FRIDAY Phase 1.} This describes the training process for the face recognizer used in Deepfake detector training. The face recognizer is trained using a closed-set and identification learning approach.}
% \label{fig:phase1}
% \end{figure}

\begin{figure}[t!]
    \centering
    \begin{subfigure}[b]{0.5\textwidth}
        \includegraphics[width=\textwidth]{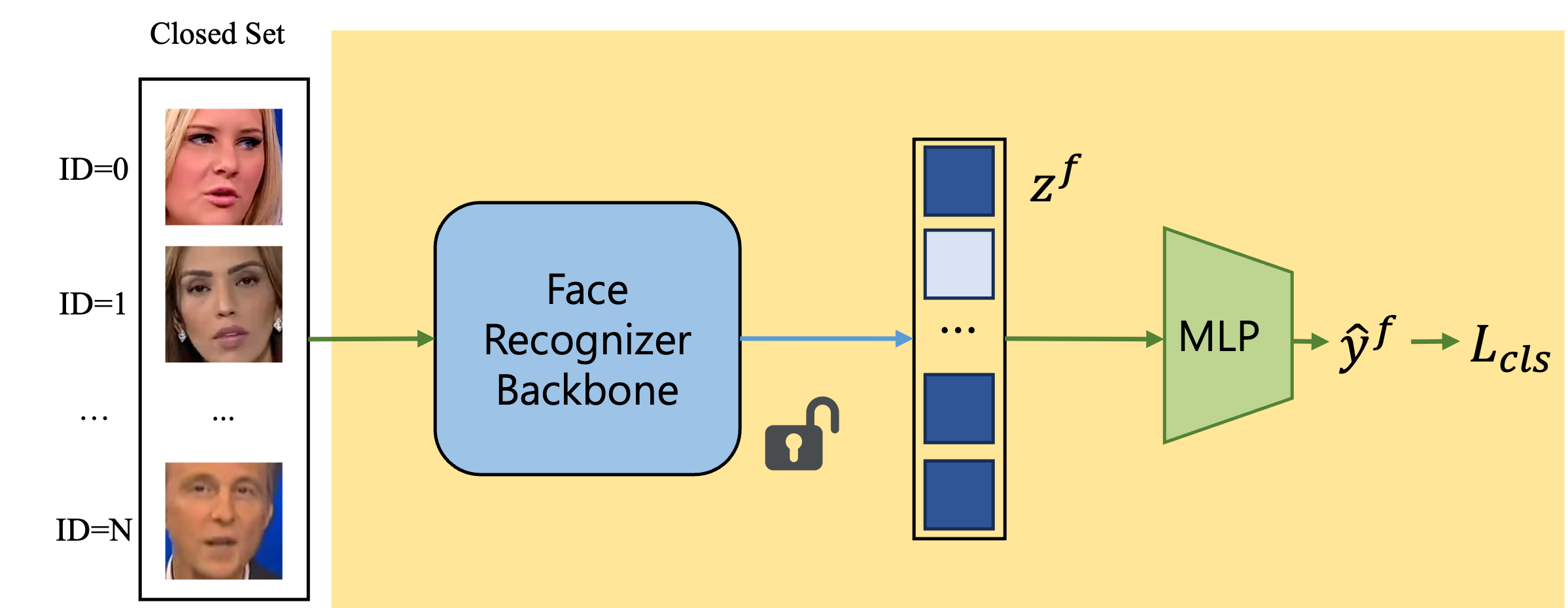}
        \caption{Phase 1: This describes the training process for the face recognizer used in Deepfake detector training. The face recognizer is trained using a closed-set and identification learning approach.}
        \label{fig:phase1}
    \end{subfigure}
    \centering
    \begin{subfigure}[b]{0.48\textwidth}
        \includegraphics[width=\textwidth]{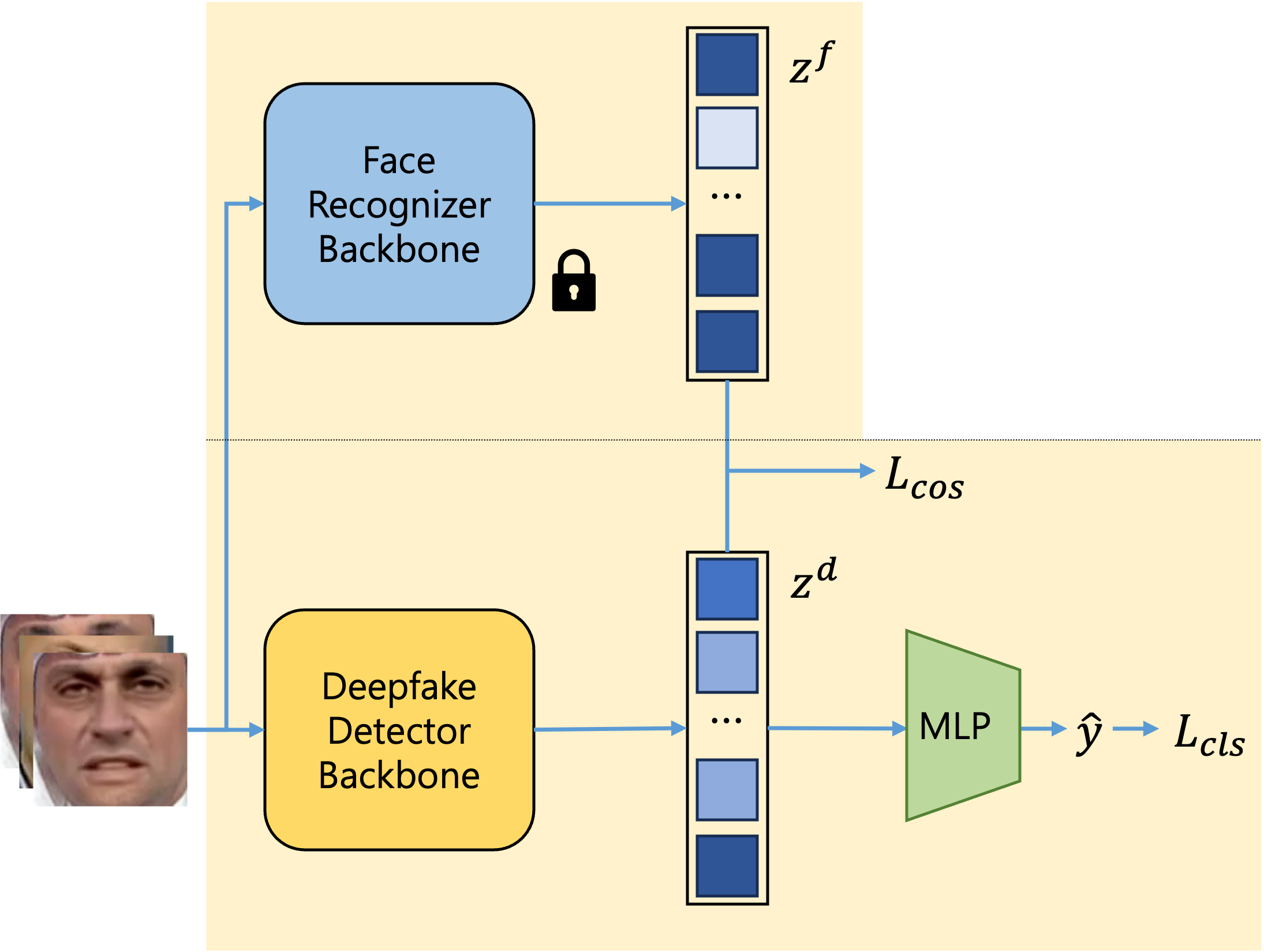}
        \caption{Phase 2: The pre-trained face recognizer from phase 1 is frozen and used to train the Deepfake detector. During this process, the embeddings are pushed apart, allowing the Deepfake detector to learn to identify Deepfakes while reducing the influence of facial identity.}
        \label{fig:phase2}
    \end{subfigure}
    \caption{\textit{FRIDAY Training Process.}}
    \label{fig:overview}
\end{figure}

% Face recognition encompasses both identification and verification, where the former identifies a person and the latter verifies if two images belong to the same person. 
We focus on face identification tasks for training the face recognizer within a closed-set framework. In addition, we only use real datasets to prevent confusion during recognizer training. This is because when Deepfake manipulation is applied, it becomes ambiguous whether to consider the person as the same individual as the original real person, which can lead to training difficulties.

% 이렇게 real datasets만 사용한 이유는 Deepfake 기술이 적용되었을 때 이미지가 real datasets의 identity와 같은 사람으로 취급시 퀄리티 차이 때문에 recognizer 학습에 혼란을 초래할 수 있기 때문이다.

The learning process is shown in Fig.~\ref{fig:phase1}. First, a classifier to distinguish \(N\) individuals is added (in this study, \(N=720\) based on the FF++ dataset). The loss function is cross entropy loss. The overall learning process can be expressed as:
\begin{equation}
z_f = f(x)
\end{equation}
\begin{equation}
\hat{y}_f = \textit{MLP}_f(z_f)
\end{equation}
\begin{equation}
L_{\text{cls}} = - \sum_{c=1}^{N} y_c \log(\hat{y}_c),
\end{equation}
where \(x \in \mathbb{R}^{B \times C \times H \times W}\) is the input original images, \(f\) is the face recognizer backbone, \(\hat{y}_f \in \mathbb{R}^{B \times N}\) is the final prediction after the classifier, $\textit{MLP}_f$, and \(y \in \mathbb{R}^{B \times N}\) is the original label used to calculate the cross entropy loss.

% \begin{figure}[]
% \centerline{\includegraphics[width=0.45\textwidth]{figures/phase2.png}}
% \caption{\textit{FRIDAY Phase 2.} 
% In phase 2, the pre-trained face recognizer from phase 1 is frozen and used to train the Deepfake detector. During this process, the embeddings are pushed apart, allowing the Deepfake detector to learn to identify Deepfakes while reducing the influence of face identity.}
% \label{fig:phase2}
% \end{figure}

\renewcommand{\arraystretch}{1.2}
\setlength{\tabcolsep}{10pt}

\begin{table*}[!t]
\centering
\caption{Performance Comparison Across Different Test Datasets (\%)}
\label{tab:main_table}
\small % Increase font size of the table
\begin{tabular}{c|cccccccc|cc}
\hline % Add top line
\multirow{4}{*}{Model} & \multicolumn{10}{c}{Test Datasets} \\ \cline{2-11} 
                       & \multicolumn{2}{c|}{In-domain} & \multicolumn{8}{c}{Cross-domain} \\ \cline{2-11}
                       & \multicolumn{2}{c|}{FF++} & \multicolumn{2}{c|}{Celeb-DF V1} & \multicolumn{2}{c|}{Celeb-DF V2} & \multicolumn{2}{c|}{DFD} & \multicolumn{2}{c}{Average} \\ \cline{2-11}
                       & ACC & \multicolumn{1}{c|}{AUC} & ACC & \multicolumn{1}{c|}{AUC} & ACC & \multicolumn{1}{c|}{AUC} & ACC & \multicolumn{1}{c|}{AUC} & ACC & \multicolumn{1}{c}{AUC} \\ \hline

CapsuleNet \cite{nguyen2019capsule}                & 89.55 & \multicolumn{1}{c|}{96.92} & 63.00 & \multicolumn{1}{c|}{69.78} & 70.66 & \multicolumn{1}{c|}{74.41} & 67.71 & \multicolumn{1}{c|}{71.67} & 67.12 & 71.95 \\
Xception \cite{rossler2019faceforensics++}              & \underline{94.47} & \multicolumn{1}{c|}{\underline{98.49}} & 64.00 & \multicolumn{1}{c|}{68.21} & 71.24 & \multicolumn{1}{c|}{72.82} & 78.20 & \multicolumn{1}{c|}{76.38} & 71.15 & 72.47 \\
CViT \cite{wodajo2021cvit}                  & 93.84 & \multicolumn{1}{c|}{98.26} & \textbf{74.00} & \multicolumn{1}{c|}{\underline{82.98}} & 74.90 & \multicolumn{1}{c|}{79.60} & 77.41 & \multicolumn{1}{c|}{78.93} & 75.44 & 80.50 \\
UIA-ViT \cite{uia}               & 93.57 & \multicolumn{1}{c|}{98.46} & \textbf{74.00}& \multicolumn{1}{c|}{82.32} & \underline{75.28} & \multicolumn{1}{c|}{\underline{80.22}} & \underline{87.41} & \multicolumn{1}{c|}{\textbf{86.14}} & \underline{78.90} & \underline{82.89} \\ \hline

FRIDAY (\(\lambda=10\)) (Ours)                 & \textbf{95.18} & \multicolumn{1}{c|}{\textbf{99.18}} & \textbf{74.00} & \multicolumn{1}{c|}{\textbf{85.27}} & \textbf{76.25} & \multicolumn{1}{c|}{\textbf{83.88}} & \textbf{90.12} & \multicolumn{1}{c|}{\underline{83.95}} & \textbf{80.12} & \textbf{84.37} \\

\hline
\end{tabular}
\end{table*}

\subsection{Phase 2: Deepfake Detector Training}
The second phase of FRIDAY learning method is shown in Fig.~\ref{fig:phase2}. First, the face recognizer, \(f\), which was pre-trained in phase 1, is frozen. An input image is fed into both the frozen face recognizer and the Deepfake detector, \(d\), to extract the respective embeddings, \(z^f, z^d\):

% FRIDAY학습 기법은 Figure (b)에 나와있는대로 입니다. 우선 phase 1에서 미리 학습시켜놓은 face recognizer (F)를 freeze 시킵니다. 그 후 이미지 입력이 들어왔을 때 Deepfake detector (d)와 freeze 시킨 face recognizer (F_freeze)에 넣고 각 embedding (z^f, z_d)을 뽑습니다:

\begin{equation}
z_f = f(x)
\end{equation}
\begin{equation}
z_d = d(x).
\end{equation}

Next, Facial Identity Attenuating loss is applied between the embeddings. By taking the absolute value of the cosine similarity, the embeddings are encouraged to become orthogonal to each other during training:

\begin{equation}
L_{\text{fia}} = \left| \frac{z^f \cdot z^d}{\|z^f\|_2 \|z^d\|_2} \right|.
\end{equation}

Finally, the embedding from the Deepfake detector is passed through a classifier, $\textit{MLP}_d$, to produce the final prediction, referred to as \(\hat{y}_d\). This prediction is then used to calculate the binary cross entropy loss based on the label \(y\) (real or fake):
% 마지막으로는 Deepfake detector에서 나온 embedding (z_d)를 fake or real을 예측하는 Classifier (MLP)에 통과시켜 최종 예측인 y_hat을 만듭니다. 해당 예측은 label (y)를 기준으로 Cross Entropy Loss를 산출합니다.
\begin{equation}
\hat{y}_d = \textit{MLP}_d(z_d)
\end{equation}
\begin{equation}
L_{\text{cls}} = -[y \log(\hat{y}_d) + (1 - y) \log(1 - \hat{y}_d)]
\end{equation}

The overall loss for training the Deepfake detector is defined as:
\begin{equation}
L_{\text{total}} = L_{\text{cls}} + \lambda \cdot L_{\text{fia}}
\end{equation}
where the parameter \(\lambda\) is a weighting factor that balances the importance of the two loss components, adjusting the influence of the Facial Identity Attenuating loss relative to the classification loss.

\section{Experiments}

\subsection{Experimental Settings}
\subsubsection{Datasets}
We conducted our experiments using four widely used datasets: FaceForensics++ \cite{rossler2019faceforensics++}, Celeb-DF v1 \& v2 \cite{li2020celeb}, and DeepfakeDetection \cite{nicholas2019dfd}. First, FaceForensics++ (FF++) comprises 1,000 original images and four different generation techniques: Deepfakes, Face2Face, FaceSwap, and NeuralTextures. Second, Celeb-DF v1 \& v2 include three types of datasets: celebrity-based celeb-real, celeb-synthesis, and general youtube-real. Third, the DeepfakeDetection (DFD) dataset, created by Google, contains 3,068 Deepfake videos and 363 original videos. We trained all models using FaceForensics++ and performed cross-domain experiments with the Celeb-DF and DeepfakeDetection datasets.

\subsubsection{Evaluation Metrics}
The evaluation metrics for the model include ACC (accuracy) and AUC (Area Under Curve). ACC is the proportion of the test set that the model correctly predicted, indicating the percentage of videos correctly identified as fake. AUC measures the performance of a classification model by evaluating the trade-off between true positive rates and false positive rates across different thresholds.

\subsubsection{Evaluation Rules}
We chose video-level evaluation, using 60 frames per video. However, some videos did not reach 60 frames due to \textit{dlib}'s inability to recognize faces in certain frames. In these instances, the evaluation was conducted by averaging the available frames. 

\subsection{Implementation Details}
\subsubsection{Data Preprocessing} We employed \textit{dlib}~\cite{king2009dlib} to extract faces from video frames, following a common practice \cite{uia, shiohara2022sbi}. When multiple faces appeared in a single frame, we extracted only the largest face to ensure consistency and focus on the primary subject. For training, we used 20 frames per video, while for evaluation, we used 60 frames per video.
\subsubsection{Other Details} For FRIDAY backbone, we used EfficientNet-B3. We utilized the Adam optimizer, starting with a learning rate of 0.00003. This was gradually reduced to 0 using a cosine learning rate scheduler over 40 epochs. Our batch size was 256. For data augmentation, we applied random flips, random crops, and random rotations to real frames. For fake frames, we used the same augmentations and added an additional Gaussian blur. Furthermore, we set \(\lambda\) to 10.
% Furthermore, based on Fig. \ref{fig:4}, \(\lambda=10\) showed the best performance, so we set \(\lambda\) to 10.
% \begin{figure*}[htbp]
%     \centerline{\includegraphics[width=0.25\textwidth]{figures/FRIDAY_comp_loss.png}}
%         \caption{\textit{FRIDAY Effectiveness.}}
%     \label{fig:friday_effect}
% \end{figure*}

\subsection{Results}
% We utilized efficientnet-b3 as the backbone and trained it using our proposed FRIDAY technique. The training was conducted on the FF++ training dataset. For in-domain experiments, we used the FF++ testing set, and for cross-domain experiments, we tested on CelebDF v1, v2, and DFD datasets. To ensure a fair comparison, we only selected models with publicly available code. 
% Our efficientnet-b3 model trained with the FRIDAY technique outperformed existing state-of-the-art models in both in-domain and cross-domain evaluations. Detailed results are presented in Table \ref{tab:main_table}.
% To ensure a fair comparison, we only selected models with publicly available code. As shown in Table \ref{tab:main_table}, our EfficientNet-B3+FRIDAY outperformed the current state-of-the-art (SOTA) methods. It outerperforms UIA-ViT, the top-performing baseline, in both in-domain and cross-domain evaluations, with the exception of DFD AUC. On average, in cross-domain settings, our method improves accuracy by 1.22\%p and AUC by 1.48\%p. These results demonstrate the robust performance of our approach.

To ensure a fair comparison, we only selected models with publicly available code. As shown in Table \ref{tab:main_table}, our FRIDAY outperformed the current state-of-the-art (SOTA) methods. 
FRIDAY outperformed the second best-performing model, Xception, on in-domain datasets by 0.71\%p in ACC and 0.69\%p in AUC. Additionally, our method outperformed the best cross-domain baseline, UIA-ViT, in all cross datasets except for DFD AUC. On average, our approach achieves a 1.22\%p higher ACC and a 1.48\%p increase in AUC. These results demonstrate the robustness of our proposed technique in both in-domain and cross-domain scenarios.

% 베이스라인들중 In-domain에서 가장 성능이 좋은 Xception과 FRIDAY를 비교하였을때, in-domain 기준 ACC 기준 0.71%오르고 AUC 기준 0.69%가 오른다. 뿐만아니라 베이스라인 중 cross-domain에서 가장 성능이 좋은 UIA-ViT와 비교하였을때 DFD AUC를 제외하고는 모든 cross dataset에서 성능이 높다. 평균적으로 보았을 때 accuracy 기준 1.22%가 올랐고 AUC 기준 1.48%가 올랐다. 이는 in-domain 그리고 cross-domain 모두에서 우리가 제안한 기법의 robustness를 보여준다.

\subsection{Effect of \(\lambda\)}
\begin{figure*}[t!]
    \centering
    \begin{subfigure}[b]{0.3\textwidth}
        \includegraphics[width=\textwidth]{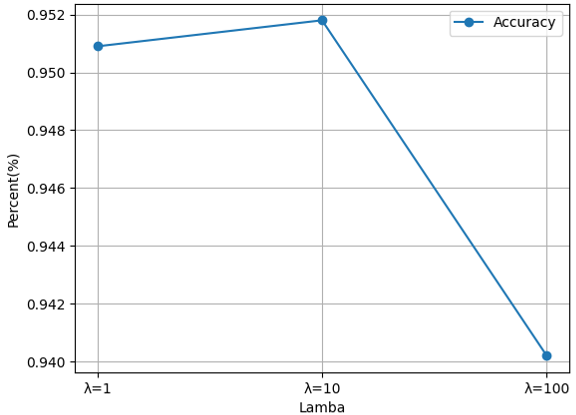}
        \caption{ACC changes along with \(\lambda\)}
        \label{fig:4a}
    \end{subfigure}
    \hfill
    \begin{subfigure}[b]{0.3\textwidth}
        \includegraphics[width=\textwidth]{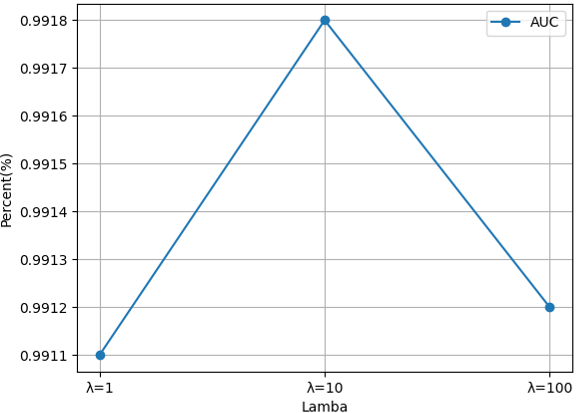}
        \caption{Lambda AUC changes}
        \label{fig:4b}
    \end{subfigure}
    \hfill
    \begin{subfigure}[b]{0.33\textwidth}
        \includegraphics[width=\textwidth]{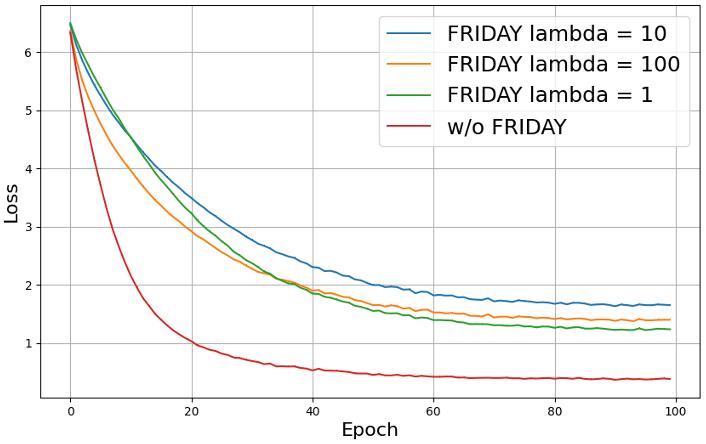}
        \caption{FRIDAY effectiveness}
        \label{fig:4c}
    \end{subfigure}
    \caption{\textit{Effect of \(\lambda\).} These figures illustrate the effect of varying \(\lambda\) on FRIDAY. (a) and (b) depict the accuracy and AUC, both of which are highest at \(\lambda=10\). (c) shows the amount of face identity using the unintentional face identity check method referenced in Fig.~s\ref{fig:face_identity_a}. According to the result, \(\lambda=10\) exhibits the lowest inclusion of face identity.}
    % (a)와 (b)는 lamda 변화에 따른 in-domain dataset에서의 accuracy와 AUC 변화량이다. 여기서 lamda=10일 때가 가장 성능이 좋다. (c)는 lamda별로 face identity check method \ref{fig:2}를 진행한 것인데 lamda=10일 때 가장 face identity 포함량이 적다.
    
    \label{fig:4}
\end{figure*}

To determine the optimal lambda for Facial Identity Attenuating loss, we conducted various comparative analyses. First, we observed the performance changes in the in-domain scenario as \(\lambda\) varied. As shown in Fig.~\ref{fig:4a} and Fig.~\ref{fig:4b}, the best AUC and ACC were achieved when \(\lambda\) was set to 10. Additionally, as depicted in Fig.~\ref{fig:face_identity_a}, we examined the changes in the facial identity content of the Deepfake detector across different lambda values. We found that when \(\lambda\) was 10, there was the least convergence, indicating a significant reduction in facial identity content. Consequently, since the reduction in facial identity content led to improved performance, we concluded that \(\lambda=10\) is the optimal value.

\section{Conclusion}
% 이 논문에서 우리는 face recognizer를 활용하여 딥페이크 탐지기 학습과정때 unintentional face identity를 atenuating 시키는 FRIDAY라는 novel 학습 기법을 제안합니다. 우리의 방식으로 학습한 모델은 기존 state-of-the-art performance를 in-domain과 cross-domain 모두에서 능가했습니다. 뿐만아니라, ablation study를 통해서 우리 기법의 효과성을 자세하게 분석하고 입증하였습니다. 하지만, 여전히 완벽하게 face identity를 감소시키지는 못하였습니다. 이는 앞으로 연구에서 더 자세히 발전되고 해결될것입니다.

We proposed a novel training technique called FRIDAY, which utilizes a face recognizer to attenuate unintentional face identity during the Deepfake detector training process. Our Facial Identity Attenuating loss forces the face recognizer embeddings and the Deepfake detector embeddings to be orthogonal. Models trained using our approach outperformed existing state-of-the-art methods in both in-domain and cross-domain settings. Additionally, we conducted analysis of \(\lambda\) to thoroughly analyze and validate the effectiveness of our technique. We hope our study inspires the use of AI to create a safer and more secure world for everyone.
% We hope our study contributes to make world safer.
% In future, we believe face identity feature will be more diminished later.

\bibliographystyle{IEEEtran}
\bibliography{ref}

\begin{thebibliography}{10}
\providecommand{\url}[1]{#1}
\csname url@samestyle\endcsname
\providecommand{\newblock}{\relax}
\providecommand{\bibinfo}[2]{#2}
\providecommand{\BIBentrySTDinterwordspacing}{\spaceskip=0pt\relax}
\providecommand{\BIBentryALTinterwordstretchfactor}{4}
\providecommand{\BIBentryALTinterwordspacing}{\spaceskip=\fontdimen2\font plus
\BIBentryALTinterwordstretchfactor\fontdimen3\font minus \fontdimen4\font\relax}
\providecommand{\BIBforeignlanguage}[2]{{%
\expandafter\ifx\csname l@#1\endcsname\relax
\typeout{** WARNING: IEEEtran.bst: No hyphenation pattern has been}%
\typeout{** loaded for the language `#1'. Using the pattern for}%
\typeout{** the default language instead.}%
\else
\language=\csname l@#1\endcsname
\fi
#2}}
\providecommand{\BIBdecl}{\relax}
\BIBdecl

\bibitem{DeepFakes}
FaceSwapDevs, ``Deepfakes,'' \url{https://github.com/deepfakes/faceswap}, 2019, accessed: 2024-07-02.

\bibitem{Face2Face}
J.~Thies, M.~Zollhofer, M.~Stamminger, C.~Theobalt, and M.~Nie{\ss}ner, ``Face2face: Real-time face capture and reenactment of rgb videos,'' in \emph{Proceedings of the IEEE conference on computer vision and pattern recognition}, 2016, pp. 2387--2395.

\bibitem{FaceSwap}
M.~Kowalski, ``Faceswap,'' \url{https://github.com/MarekKowalski/Faceswap}, 2018, accessed: 2024-07-02.

\bibitem{FaceShifter}
L.~Li, J.~Bao, H.~Yang, D.~Chen, and F.~Wen, ``Faceshifter: Towards high fidelity and occlusion aware face swapping,'' \emph{arXiv preprint arXiv:1912.13457}, 2019.

\bibitem{rossler2019faceforensics++}
A.~Rossler, D.~Cozzolino, L.~Verdoliva, C.~Riess, J.~Thies, and M.~Nie{\ss}ner, ``Faceforensics++: Learning to detect manipulated facial images,'' in \emph{Proceedings of the IEEE/CVF international conference on computer vision}, 2019, pp. 1--11.

\bibitem{afchar2018mesonet}
D.~Afchar, V.~Nozick, J.~Yamagishi, and I.~Echizen, ``Mesonet: a compact facial video forgery detection network,'' in \emph{2018 IEEE international workshop on information forensics and security (WIFS)}.\hskip 1em plus 0.5em minus 0.4em\relax IEEE, 2018, pp. 1--7.

\bibitem{nguyen2019capsule}
H.~H. Nguyen, J.~Yamagishi, and I.~Echizen, ``Use of a capsule network to detect fake images and videos,'' \emph{arXiv preprint arXiv:1910.12467}, 2019.

\bibitem{noise_based}
T.~Wang and K.~P. Chow, ``Noise based deepfake detection via multi-head relative-interaction,'' in \emph{Proceedings of the AAAI Conference on Artificial Intelligence}, vol.~37, no.~12, 2023, pp. 14\,548--14\,556.

\bibitem{uia}
W.~Zhuang, Q.~Chu, Z.~Tan, Q.~Liu, H.~Yuan, C.~Miao, Z.~Luo, and N.~Yu, ``Uia-vit: Unsupervised inconsistency-aware method based on vision transformer for face forgery detection,'' in \emph{European conference on computer vision}.\hskip 1em plus 0.5em minus 0.4em\relax Springer, 2022, pp. 391--407.

\bibitem{li2020facexray}
L.~Li, J.~Bao, T.~Zhang, H.~Yang, D.~Chen, F.~Wen, and B.~Guo, ``Face x-ray for more general face forgery detection,'' in \emph{Proceedings of the IEEE/CVF conference on computer vision and pattern recognition}, 2020, pp. 5001--5010.

\bibitem{li2018exposing}
Y.~Li and S.~Lyu, ``Exposing deepfake videos by detecting face warping artifacts,'' \emph{arXiv preprint arXiv:1811.00656}, 2018.

\bibitem{shiohara2022sbi}
K.~Shiohara and T.~Yamasaki, ``Detecting deepfakes with self-blended images,'' in \emph{Proceedings of the IEEE/CVF Conference on Computer Vision and Pattern Recognition}, 2022, pp. 18\,720--18\,729.

\bibitem{implicit_identity_leakage}
S.~Dong, J.~Wang, R.~Ji, J.~Liang, H.~Fan, and Z.~Ge, ``Implicit identity leakage: The stumbling block to improving deepfake detection generalization,'' in \emph{Proceedings of the IEEE/CVF Conference on Computer Vision and Pattern Recognition}, 2023, pp. 3994--4004.

\bibitem{dong2022celebrities}
X.~Dong, J.~Bao, D.~Chen, T.~Zhang, W.~Zhang, N.~Yu, D.~Chen, F.~Wen, and B.~Guo, ``Protecting celebrities from deepfake with identity consistency transformer,'' in \emph{Proceedings of the IEEE/CVF Conference on Computer Vision and Pattern Recognition}, 2022, pp. 9468--9478.

\bibitem{wodajo2021cvit}
D.~Wodajo and S.~Atnafu, ``Deepfake video detection using convolutional vision transformer,'' \emph{arXiv preprint arXiv:2102.11126}, 2021.

\bibitem{li2020celeb}
Y.~Li, X.~Yang, P.~Sun, H.~Qi, and S.~Lyu, ``Celeb-df: A large-scale challenging dataset for deepfake forensics,'' in \emph{Proceedings of the IEEE/CVF conference on computer vision and pattern recognition}, 2020, pp. 3207--3216.

\bibitem{nicholas2019dfd}
D.~Nicholas, G.~Andrew, K.~Per, V.~V. Alexey, L.~Thomas, C.~Jeremiah, and B.~Christoph, ``Deepfakes detection dataset by google \& jigsaw,'' 2019, available at: \url{https://ai.googleblog.com/2019/09/contributing-data-to-deepfake-detection.html}.

\bibitem{king2009dlib}
D.~E. King, ``Dlib-ml: A machine learning toolkit,'' \emph{The Journal of Machine Learning Research}, vol.~10, pp. 1755--1758, 2009.

\end{thebibliography}

\end{document}